\RequirePackage{amsmath, accents}
\documentclass[runningheads]{llncs}
\usepackage[T1]{fontenc}
\usepackage{graphicx}

\usepackage{amssymb}   
\usepackage{mathtools} 
\usepackage{mathrsfs}  
\usepackage{enumitem}  
\usepackage{dsfont}    
\usepackage{stmaryrd}  
\SetSymbolFont{stmry}{bold}{U}{stmry}{m}{n}

\usepackage{graphicx}  
\usepackage{adjustbox}
\graphicspath{{figurer/}}
\usepackage{booktabs, multirow}  
 \usepackage[table]{xcolor} 

\usepackage{accents}
\newcommand{\thickbar}{} 
\DeclareRobustCommand*\thickbar[1]{\accentset{\rule{.35em}{.65pt}}{#1}}

\DeclarePairedDelimiter{\ceil}{\lceil}{\rceil}
\DeclarePairedDelimiter{\floor}{\lfloor}{\rfloor}

\newcommand{\set}[1]{\left\lbrace #1 \right\rbrace}
\newcommand{\abs}[1]{\left| #1 \right|}%
\newcommand{\tm}[1]{\texttt{#1}}%

\newcommand{\diff}{\mathop{}\!\mathrm{d}}

\newcommand{\R}{\mathbb{R}}   
\newcommand{\D}{{\mathcal{D}}}
\newcommand{\E}{\mathbb{E}}   

\newcommand{\w}{\boldsymbol{w}}
\newcommand{\W}{\boldsymbol{W}}
\newcommand{\x}{\boldsymbol{x}}
\newcommand{\y}{\boldsymbol{y}}

\newcommand{\xsb}{{\boldsymbol{x}_{\thickbar{\mathcal{S}}}}}

\newcommand{\xs}{{\boldsymbol{x}_{\mathcal{S}}}}
\newcommand{\xss}{{\boldsymbol{x}_{\mathcal{S}}^*}}
\newcommand{\sbb}{{\thickbar{\mathcal{S}}}}
\newcommand{\s}{{\mathcal{S}}}
\newcommand{\M}{{\mathcal{M}}}
\newcommand{\pow}{{\mathcal{P}}}

\newcommand{\bphi}{{\boldsymbol{\phi}}}
\newcommand{\bmu}{{\boldsymbol{\mu}}}

\newcommand{\bbeta}{{\boldsymbol{\beta}}}

\newcommand{\bSigma}{{\boldsymbol{\Sigma}}}

\newcommand{\gaussian}{\texttt{gaussian}}

\newcommand{\shapr}{\texttt{shapr}}

\DeclareMathOperator*{\argmin}{arg\,min}

\newcommand{\unique}{\texttt{unique}}
\newcommand{\paired}{\texttt{paired}}

\newcommand{\pairedckernel}{\texttt{paired c-kernel}}

\newcommand{\Unique}{\texttt{Unique}}
\newcommand{\Paired}{\texttt{Paired}}

\newcommand{\Pairedckernel}{\texttt{Paired c-kernel}}

\newcommand{\pyshap}{\texttt{PySHAP}}
\newcommand{\pyshaps}{\texttt{PySHAP\textsuperscript{*}}}

\newcommand{\pyshapsckernel}{\texttt{PySHAP\textsuperscript{*}\,c-kernel}}

\usepackage[bottom]{footmisc}
\usepackage{makecell} 
\newcommand*{\rom}[1]{\expandafter\@slowromancap\romannumeral #1@} 

\usepackage{pifont}  
\newcommand{\cmark}{\textcolor{green}{\ding{51}}} 
\newcommand{\xmark}{\textcolor{red}{\ding{55}}}  
\newcommand{\pmark}{\textcolor{orange}{$\boldsymbol{\sim}$}} 

\usepackage{varioref}
\usepackage[pdfusetitle]{hyperref}
\usepackage[nameinlink, noabbrev]{cleveref}
\usepackage{color}

\begin{document}

\title{Improving the Weighting Strategy in KernelSHAP}

\author{Lars Henry Berge Olsen\texorpdfstring{\inst{1,2}\orcidID{0009-0006-9360-6993}}{} \\ \and
Martin Jullum\texorpdfstring{\inst{2}\orcidID{0000-0003-3908-5155}}{}}

\authorrunning{Olsen and Jullum}

\institute{Department of Mathematics, University of Oslo, Norway \and
Norwegian Computing Center, Norway \email{\{lhbolsen,jullum\}@nr.no}}

\maketitle              

\begin{abstract}
In Explainable AI (XAI), Shapley values are a popular model-agnostic framework for explaining predictions made by complex machine learning models. The computation of Shapley values requires estimating non-trivial \textit{contribution functions} representing predictions with only a subset of the features present. As the number of these terms grows exponentially with the number of features, computational costs escalate rapidly, creating a pressing need for efficient and accurate approximation methods. For tabular data, the \texttt{KernelSHAP} framework is considered the state-of-the-art model-agnostic approximation framework. \texttt{KernelSHAP} approximates the Shapley values using a weighted sample of the contribution functions for different feature subsets. We propose a novel modification of \texttt{KernelSHAP} which replaces the stochastic weights with deterministic ones to reduce the variance of the resulting Shapley value approximations. This may also be combined with our simple, yet effective modification to the \texttt{KernelSHAP} variant implemented in the popular Python library \texttt{SHAP}. Additionally, we provide an overview of established methods. Numerical experiments demonstrate that our methods can reduce the required number of contribution function evaluations by $5\%$ to $50\%$ while preserving the same accuracy of the approximated Shapley values -- essentially reducing the running time by up to $50\%$. These computational advancements push the boundaries of the feature dimensionality and number of predictions that can be accurately explained with Shapley values within a feasible runtime.

\end{abstract}

\section{Introduction}
\label{Introduction}
The field of Explainable Artificial Intelligence (XAI) has developed various explanation frameworks to provide insights into the inner workings of complex models and to make their predictions more understandable to humans \cite{adadi2018peeking,covert2021explaining,molnar2019}. Developing and adopting XAI frameworks is crucial for bridging the gap between model complexity and transparency, trustworthiness, and explainability. 
One of the most used explanation frameworks is \textit{Shapley values} \cite{molnar2023SHAP,shapley1953value}. 

Shapley values stem from cooperative game theory but are in the context of XAI, often used as a \textit{local} feature attribution framework that explains how the features contribute to the prediction $f(\x)$ made by a complex predictive model $f$. Local means that we explain the prediction of single observations $\x$.

Computing Shapley values is generally an NP-hard problem \cite{deng1994complexity,faigle1992shapley}, and directly calculating Shapley values for feature attribution has a computational complexity which is exponential in the number of features $M$. 
The reason for this is that the Shapley value formula requires evaluating a \textit{contribution function} $v(\s)$ for every subset (coalition) $\s$ of the set of features, which in total amounts to $2^M$ different feature subsets.
Evaluation of $v(\s)$ is typically costly, especially for \textit{conditional} Shapley values, which is often preferable over \textit{marginal} Shapley values for real life model explanations \cite{chen2020true,covert2021explaining}. 
Reducing the number of evaluations of $v(\s)$ for different coalitions $\s$ therefore directly reduces the computational cost, which is crucial for making (conditional) Shapley values a feasible model-agnostic explanation framework beyond low and medium dimensional feature spaces.

A variety of approximation strategies have been proposed in the literature \cite{chen2023algorithms}. In this paper, we consider the widely used, model-agnostic approximation method \texttt{KernelSHAP} \cite{lundberg2017unified}, where model-agnostic means that it is applicable to any model $f$. \texttt{KernelSHAP} formulates the Shapley values as the solution to a certain weighted least squares problem, and then approximates that solution by using a weighted sample of the coalitions $\s$, instead of all the $2^{M}$ combinations. It is widely recognized as the state-of-the-art model-agnostic approximation framework for tabular data \cite{chen2022algorithms,kolpaczki2024approximating}, largely due to its broad utilization of the $v(\s)$ evaluations. In its original form, \texttt{KernelSHAP} sample coalitions $\s$ with replacement, and weigh the elements in the unique set of coalitions based on their sampling frequency. A drawback of this procedure is the random and undesirable variability in the weights of the sampled coalitions, which is caused by the stochastic sampling frequency.

The main contribution of the paper is three-fold. First, we provide an overview of established sampling and weighting strategies within \texttt{KernelSHAP}. Second, we propose a modification of the \texttt{KernelSHAP} procedure, where we for a given sample of coalitions, replace the stochastic weights with deterministic weights. This useful modification reduces the variance of the resulting Shapley value approximations. We incorporate our deterministic weighting procedure not only into the original \texttt{KernelSHAP} procedure, but also into various existing extensions and improvements of the original procedure, such as antithetic/paired sampling and semi-deterministic sampling. Additionally, we propose a simple yet effective modification to the variant of \texttt{KernelSHAP} implemented in the widely-used \href{https://shap.readthedocs.io/en/latest/}{\texttt{SHAP}} Python library \cite{lundberg2017unified}. Third, through simulation studies with tabular data, we compare the strategies and demonstrate that our \texttt{KernelSHAP} modifications consistently outperform existing methods. Specifically, we illustrate how our best-performing procedure reduces the required number of coalitions by $5-50\%$, potentially halving the running time, compared to the \texttt{KernelSHAP} method implemented in the \href{https://shap.readthedocs.io/en/latest/}{\texttt{SHAP}} Python library, while maintaining the same accuracy.

Finally, note that all our modified \texttt{KernelSHAP} procedures are perfectly applicable to other 
Shapley value applications within XAI, such as \textit{global} feature attributions (i.e., the overall model performance) \cite{covert2020understanding} and data valuation \cite{ghorbani2019data}, and frankly any other application of Shapley values also outside of the XAI domain. Investigating the efficiency in such situations is, however, outside the scope of this paper.

\Cref{CSVE} introduces the Shapley value explanation framework, and describe how to compute them exactly and approximately using the \texttt{KernelSHAP} framework. \Cref{Sampling_strategies} outlines coalition sampling and weighting strategies within the \texttt{KernelSHAP} framework, encompassing both established methods and our novel approaches.
\Cref{Simulations} presents numerical simulation studies comparing the accuracy of the different approximation strategies, while \Cref{sec:real_world_data} conducts experiments on real-world data. Finally, \Cref{Conclusion} provides conclusions and outlines further work.

\vspace{0.5ex}
\section{Shapley Values} \label{CSVE}  \vspace{0.5ex}
Originally, Shapley values were proposed as a solution concept for how to divide the payout of a cooperative game $v:\mathcal{P}(\M) \mapsto \R$ onto the players based on four axioms \cite{shapley1953value}. The game is played by $M$ players where $\M = \{1,2,\dots,M\}$ denotes the set of all players and $\mathcal{P}(\M)$ is the power set, that is, the set of all subsets of $\M$. We call $v(\s)$ the \textit{contribution function}\footnote{It is also called the \textit{value}, \textit{reward}, \textit{lift}, and \textit{characteristic} function in the literature.} and it maps a subset of players $\s \in \pow(\M)$, also called a \textit{coalition}, to a real number representing their contribution in the game $v$. The Shapley values $\phi_j = \phi_j(v)$ assigned to each player $j$, for $j = 1, \dots, M$, uniquely satisfy the following properties:
\begin{enumerate}[align=left, itemsep=2pt, topsep=6pt] 
    \item [\textbf{Efficiency}:] They sum to the value of the grand coalition $\M$ minus the empty set $\emptyset$, that is, $\sum_{j=1}^M \phi_j = v(\M) - v(\emptyset)$.
    \item [\textbf{Symmetry}:] Two equally contributing players $j$ and $k$, that is,  $v(\s \cup \{j\}) = v(\s \cup \{k\})$ for all $\s$, receive equal payouts $\phi_j = \phi_k$.
    \item [\textbf{Dummy}:] A non-contributing player $j$, that is, $v(\s) = v(\s \cup \{j\})$ for all $\s$, receives $\phi_j = 0$.
    \item [\textbf{Linearity}:] A linear combination of $n$ games $\{v_1, \dots, v_n\}$, that is, \newline $v(\s) = \sum_{k=1}^nc_kv_k(\s)$, has Shapley values given by $\phi_j(v) = \sum_{k=1}^nc_k\phi_j(v_k)$.
\end{enumerate}
    
The values $\phi_j, j = 1,\ldots,M$, which uniquely satisfy these axioms, were shown by \cite{shapley1953value} to be given by the formula:
\begin{align}
    \label{eq:ShapleyValuesDef}
    \phi_j = \sum_{\s \in \pow(\M \backslash \{j\})} \frac{|\s|!(M-|\s|-1)!}{M!}\left(v(\s \cup \{j\}) - v(\s) \right),
\end{align}
where $|\mathcal{S}|$ is the number of players in coalition $\s$. The number of terms in \eqref{eq:ShapleyValuesDef} is $2^{M}$, hence, the complexity grows exponentially with the number of players $M$. 

When using Shapley values for local feature attributions of a prediction model, the cooperative game $v(\s)$ is related to the predictive model $f$, the features represent the players, and the payout is the predicted response $f(\x)$, for a specific explicand $\x = \x^*$. The Shapley value $\phi_j$ in \eqref{eq:ShapleyValuesDef} is then a weighted average of the $j$th feature’s marginal contribution to each coalition $\mathcal{S}$ and describes the importance of the $j$th feature in the prediction $f(\boldsymbol{x}^*) = \phi_0 + \sum_{j=1}^M\phi_j^*$, where $\phi_0$ denotes the value not assigned to any of the features, typically set to the mean prediction $\E\left[f(\boldsymbol{x})\right]$. That is, the Shapley values sums to the difference between the prediction $f(\boldsymbol{x}^*)$ and the global mean prediction. 

In this paper, we consider the tabular data setting of supervised learning where the predictive model $f(\x)$ is trained on $\mathcal{X} = \{\boldsymbol{x}^{[i]}, y^{[i]}\}_{i = 1}^{N_\text{train}}$. Here $\boldsymbol{x}^{[i]}$ is an $M$-dimensional feature vector, $y^{[i]}$ is a univariate response, and $N_\text{train}$ is the number of training observations. Additionally, we focus on \textit{conditional Shapley values}, which is properly defined below. Conditional Shapley values incorporates feature dependencies into the explanations \cite{aas2019explaining}, contrasting \textit{marginal Shapley values} where one omits the dependencies \cite{chen2023algorithms}. The two versions differ only in their definition of $v(\s)$ and coincide when the features are independent. Conditional Shapley values, introduced below, may be preferable in many situations \cite{aas2019explaining,chen2020true,covert2021explaining}, and are consistent with standard probability axioms \cite[Proposition 7]{covert2021explaining}. However, they come at a higher computational cost due to the need to model dependencies across arbitrary feature subsets.  This computational burden likely explains why most software implementations, and to some extent XAI-research, are oriented around the marginal approach, further limiting the practical adoption of conditional Shapley values. We focus on the conditional approach since that is where efficiency enhancements are most critical. However, all methods in this paper apply equally to marginal Shapley values, where computational efficiency is also a significant concern. Throughout this article, we refer to conditional Shapley values when discussing Shapley values unless otherwise specified.

For conditional Shapley value explanation framework, the contribution function $v(\s)$ in \eqref{eq:ShapleyValuesDef} is the expected response of $f(\boldsymbol{x})$ conditioned on the features in $\mathcal{S}$ taking on the values $\boldsymbol{x}_\mathcal{S}^*$ \cite{lundberg2017unified}. That is, for continuously distributed features,
\begin{align}
    \label{eq:ContributionFunc}
    \begin{split}
        v(\s) 
        &=
        \E\big[ f( \underbrace{\xsb, \xs}_{\x}) | \xs = \xss \big] 
        = 
        \int f(\xsb, \xss) p(\xsb | \xs = \xss) \diff \xsb,
    \end{split}
\end{align}
with an equivalent formula using sums for discretely distributed features.
Here, $\xs = \{x_j:j \in \mathcal{S}\}$ denotes the features in subset $\mathcal{S}$, $\xsb = \{x_j:j \in \thickbar{\mathcal{S}}\}$ denotes the features outside $\mathcal{S}$, that is, $\thickbar{\mathcal{S}} = \mathcal{M}\backslash\mathcal{S}$, and $p(\xsb | \xs = \xss)$ is the conditional density of $\xsb$ given $\xs = \xss$. 
To compute the Shapley values in \eqref{eq:ShapleyValuesDef}, we need to compute \eqref{eq:ContributionFunc} for all $\s \in \pow(\M)$, except for the edge cases $\mathcal{S} \in \set{\emptyset, \M}$. For $\s = \emptyset$, we have by definition that $\phi_0 = v(\emptyset) = \E[f(\x)]$, where the average training response $\overline{y}_\text{train}$ is a commonly used estimate \cite{aas2019explaining}. Moreover, for $\s = \M$, we have $\xs = \x^*$ and $v(\mathcal{M}) = f(\x^*)$. We denote the non-trivial coalitions by $\pow^*(\M) = \pow(\M) \backslash \{\emptyset, \M\}$. 

Computing \eqref{eq:ContributionFunc} is not straightforward for a general data distribution and model. Assuming independent features, or having $f$ be linear, simplifies the computations \cite{aas2019explaining,lundberg2017unified}, but these assumptions do not generally hold. There is a wide range of methods used to estimate $v(\s)$, using, e.g., Gaussian assumptions \cite{aas2019explaining,chen2020true}, conditional inference trees \cite{redelmeier2020}, or variational auto-encoders \cite{Olsen2022}. See \cite{olsen2024comparative} for an extensive overview.

\subsection{Approximation Strategies}
\label{subsubsec:ShapleyValuesExplainability:ApproximationStrategies}
In this section, we highlight procedures that use approximations or model assumptions to reduce the computational complexity of the Shapley value explanation framework and make the computations tractable in higher dimensions. The approximative speed-up strategies can be divided into model-specific and model-agnostic strategies \cite{chen2023algorithms}.

The model-specific strategies put assumptions on the predictive model $f$ to improve the computational cost, but some of the strategies are restricted to marginal Shapley values. For conditional Shapley values, \cite{aas2019explaining,chen2020true} derive explicit expressions for linear models to speed up the computations, and \cite{lundberg2020local} proposes the path-dependent \texttt{TreeSHAP} algorithm for tree-based models. \cite{yang2021fast} improves the speed of the \texttt{TreeSHAP} algorithm by pre-computing expensive steps at the cost of a slightly higher memory consumption, and the \texttt{Linear TreeSHAP} \cite{bifet2022linear} reduces the time complexity from polynomial to linear time. 
There are also speed-up strategies for deep neural network models, but they are limited to marginal Shapley values \cite{ancona2019explaining,wang2020shapley}.

The model-agnostic strategies put no assumptions on the predictive model $f$ and often use stochastic sampling-based estimators \cite{aas2019explaining,castro2017improving,lundberg2017unified,mitchell2022sampling,okhrati2021multilinear}. That is, to speed up the computations, they approximate the Shapley value explanations by a sampled subset of the coalitions instead of considering all of them. Thus, the strategies are stochastic, but converge to the exact solution. One of the most common model-agnostic strategies is the aforementioned \texttt{KernelSHAP} strategy \cite{covert2021improving,lundberg2017unified}, which we consider and introduce properly in Section \ref{seq:kernelSHAP}. 

There exist other approximation strategies linked to alternative Shapley formulations, such as the permutation sampling-based frameworks based on the random order value formulation \cite{chen2023algorithms,strumbelj2010efficient,strumbelj2014explaining}. We focus on improving the \texttt{KernelSHAP} framework as it has superior efficiency compared to permutation sampling-based frameworks. In the context of conditional Shapley values, the primary computational burden is related to the number of $v(\s)$ values that need to be calculated. \texttt{KernelSHAP} leverages \textit{all} computed $v(\s)$ values to determine the Shapley values for \textit{all} features. In contrast, permutation sampling only uses a portion of the evaluated $v(\s)$ values to estimate each feature's Shapley value, resulting in less efficient utilization of these values. Examples of approximation strategies for the permutation sampling-based framework include orthogonal spherical codes \cite{mitchell2022sampling}, stratified sampling \cite{castro2017improving,maleki2015addressing}, and many others \cite{covert2021improving,illes2019estimation,mitchell2022sampling}. However, these strategies typically cannot be directly integrated into the \texttt{KernelSHAP} approximation framework and are thus considered outside the scope of this article. Another method by \cite{kolpaczki2024approximating}, based on splitting the summand in \eqref{eq:ShapleyValuesDef} into two components to be estimated separately, also utilizes all computed $v(\s)$ for all Shapley value. In their simulation experiments for local model explanations, it performed similar or worse than an unpaired version of \texttt{KernelSHAP}. For a broad introduction to various model-agnostic and model-specific strategies approximation strategies for Shapley values in XAI, we refer to \cite{chen2023algorithms}.

\vspace{-0.5ex} \subsection{The \texttt{KernelSHAP} Framework}\vspace{-0.5ex}
\label{seq:kernelSHAP}

\cite{lundberg2017unified} shows that the Shapley value formula in \eqref{eq:ShapleyValuesDef} may also be conveniently expressed as the solution of the following weighted least squares problem: 
\begin{align}
    \label{eq:ShapleyValuesDefWLS}
     \argmin_{\bphi \in \R^{M+1}}\sum_{\s \in \pow(\M)} k(M, |\s|)\Big(\phi_0 + \sum_{j \in \s}\phi_j - v(\s)\Big)^2,    
\end{align}
where 
\begin{align}
\label{eq:ShapleyKernelWeights}
    k(M, |\s|) = \frac{M-1}{\binom{M}{|\s|}|\s|(M-|\s|)},
\end{align}
for $|\s| = 0,1,2,\dots,M$, are the \textit{Shapley kernel weights} \cite{charnes1988extremal,lundberg2017unified}. In practice, the infinite Shapley kernel weights $k(M,0) = k(M, M) = \infty$ can be set to a large constant $C = 10^6$ \cite{aas2019explaining}. The matrix solution of \eqref{eq:ShapleyValuesDefWLS} is 
\begin{align}
    \label{eq:ShapleyValuesDefWLSSolution}
    \boldsymbol{\phi} =(\boldsymbol{Z}^T\boldsymbol{W}\boldsymbol{Z})^{-1}\boldsymbol{Z}^T\boldsymbol{W}\boldsymbol{v} = \boldsymbol{R}\boldsymbol{v}.
\end{align}
Here $\boldsymbol{Z}$ is a $2^M \times (M+1)$ matrix with $1$s in the first column (to obtain $\phi_0$) and the binary representations\footnote{For example, the binary representation of $\s = \{1,3\}$ when $M=4$ is $[1,0,1,0]$.} of the coalitions $\s \subseteq \M$ in the remaining columns. While $\boldsymbol{W} = \operatorname{diag}(C, \w, C)$ is a $2^M \times 2^M$ diagonal matrix containing the Shapley kernel weights $k(M, |\s|)$. The $\w$ vector contains the $2^M-2$ finite Shapley kernel weights, which we normalize to sum to one for numerical stability. Finally, $\boldsymbol{v}$ is a column vector of height $2^M$ containing the contribution function values $v(\s)$. The $\s$ in $\boldsymbol{W}$ and $\boldsymbol{v}$ corresponds to the coalition of the corresponding row in $\boldsymbol{Z}$. The $\boldsymbol{R}$ matrix is independent of the explicands. When explaining $N_\text{explain}$ predictions, we can replace $\boldsymbol{v}$ with a $2^M \times N_\text{explain}$ matrix $\boldsymbol{V}$, where column $i$ contains the contribution functions for the $i$th explicand.

The weighted least squares solution formulation naturally motivates approximate solutions by solving \eqref{eq:ShapleyValuesDefWLS} using sampled subset of coalitions $\D \subseteq \pow(\M)$ (with replacement) instead of all coalitions $\s \in \pow(\M)$. 
This is the \texttt{KernelSHAP} approximation framework \cite{lundberg2017unified} and the corresponding approximation is
\begin{align}
    \label{eq:ShapleyValuesDefWLSSolution_approx}
    \boldsymbol{\phi}_{\D} =(\boldsymbol{Z}_{\D}^T\boldsymbol{W}_{\D}\boldsymbol{Z}_{\D})^{-1}\boldsymbol{Z}_{\D}^T\boldsymbol{W}_{\D}\boldsymbol{v}_{\D}, 
\end{align}
where only the $N_\text{coal} = |\D|$ unique coalitions in $\D$ are used. If a coalition $\s$ is sampled $K$ times, then the corresponding weight in $\W_\D = \operatorname{diag}(C, \w_\D, C)$, denoted by $w_\s$, is proportional to $K$, as we normalize the weights for numerical stability. The \texttt{KernelSHAP} framework is also useful in lower dimensions if $v(\s)$ is expensive to compute. \cite{williamson2020efficient} shows that the \texttt{KernelSHAP} approximation framework is consistent and asymptotically unbiased, while \cite{covert2021improving} shows that it is empirically unbiased for even a modest number of coalitions. A particularly nice property of \texttt{KernelSHAP} is that the full $\boldsymbol{v}_{\D}$ is utilized to estimate \textit{all} the $M$ Shapley values, making it a sampling efficient approximation method.

\vspace{-0.5ex}
\section{Sampling and Weighting Strategies} \vspace{-0.5ex}
\label{Sampling_strategies}
In this section, we describe established (\unique, \paired, and \pyshap) and novel (\pairedckernel, \pyshaps and \pyshapsckernel) strategies for selecting the $N_\text{coal}$ unique coalitions in $\D$ and how to weigh them in \eqref{eq:ShapleyValuesDefWLSSolution_approx} when approximating Shapley value explanations. \Cref{tab:Strategies_summary} gives an overview of all the strategies and their characteristic properties. The empty and grand coalitions are always included in the approximations; thus, they are excluded from the sampling procedure. Consequently, $N_\text{coal}$ is an integer between (exclusive) $2$ and $2^M$, as the full set of coalitions yields exact Shapley values. To simplify our descriptions and derivations below, let $T_{L}(\s)$ denote the number of times coalition $\s$ has been sampled, after drawing $L$ coalitions (resulting in the $N_{\s}$ unique coalitions).

\begin{figure}[t]
    \centering
    \centerline{\includegraphics[width=1\textwidth]{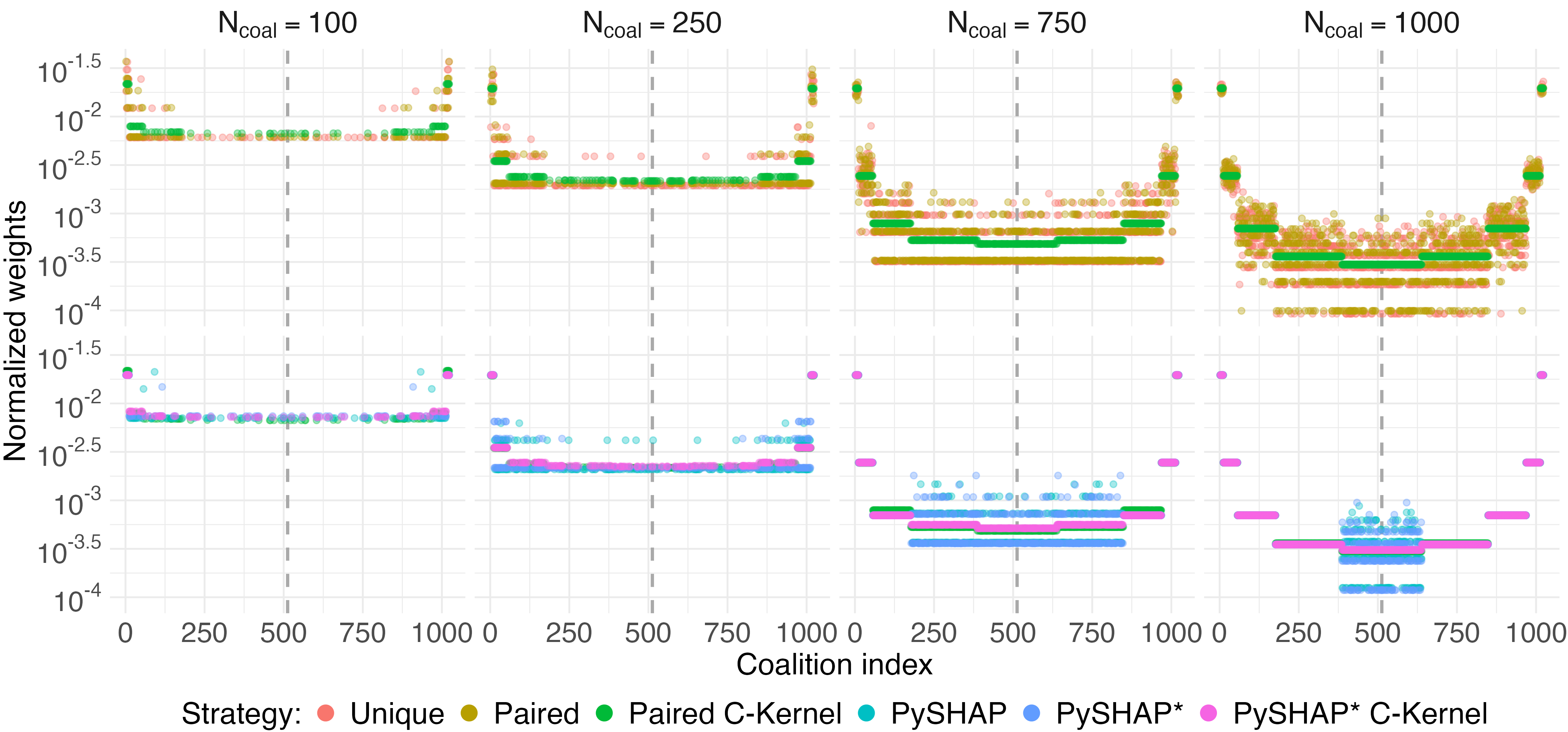}}
    \vspace{-1ex}
    \caption{{\small The normalized weights $w_\s$ used in \eqref{eq:ShapleyValuesDefWLSSolution_approx} by the strategies in \Cref{Sampling_strategies} for different number of unique coalitions $N_\text{coal}$ in an $M = 10$-dimensional setting. The \texttt{paired}-based strategies are symmetric around the vertical line. The \pairedckernel\ and \pyshapsckernel\ strategies both have identical weights within each coalition size, but their weights are slightly different from each other.
    }}    
    \vspace{-0.5ex}
    \label{fig:Sampling}
\end{figure}

In \Cref{fig:Sampling}, we illustrate the normalized weights $w_\s$ used in \eqref{eq:ShapleyValuesDefWLSSolution_approx} by the different strategies introduced below for an $M=10$-dimensional setting with $N_\text{coal} \in \{100, 250, 750, 1000\}$. We index the coalitions by first ordering them based on coalition size, i.e., $\{1\}$ precedes $\{1,2\}$, and then by their elements for equal-sized coalitions, i.e., $\{1,2\}$ precedes $\{1,3\}$. This indexing ensures that coalitions with indices $i$ and $2^M + 1 - i$ are complementary, for $i = 1, 2, \dots, 2^M$, as seen by the paired strategies where the weights are symmetric around the dashed vertical lines in \Cref{fig:Sampling}. The weights of the empty ($i = 1$) and grand ($i = 2^M = 1024$) coalitions are omitted as they are strategy-independent and infinite.

\renewcommand{\arraystretch}{1.5} 
\begin{table}[t]
  \centering
  \rowcolors{2}{gray!20}{white}
  \centerline{
    \resizebox{1\textwidth}{!}{
    \begin{tabular}{lccccccc}
      \specialrule{.8pt}{0pt}{2pt}
      Strategy & \makecell{Stochastic\\sampling} & \makecell{Paired\\sampling} & \makecell{Equal weights\\within each\\coalition size} & \makecell{Weight $w_\s$\\converges\\to $p_\s$} & \makecell{Weight $w_\s$ is proportional to} \\ 
      \specialrule{.4pt}{2pt}{0pt}
        \hyperref[sampling:unique]{\Unique} & \cmark & \xmark & \xmark & \pmark & $T_{L}(\s)$ \\
        \hyperref[sampling:paired]{\Paired} & \cmark & \cmark  & \xmark & \pmark & $T_{L}(\s) + T_{L}(\sbb)$ \\
        \hyperref[sampling:paired_c_kernel]{\textcolor{blue}{\Pairedckernel}} & \cmark & \cmark & \cmark & \cmark & $ 2p_\s \big/ (1-(1-2p_\s)^{L/2})$ \\
        \hyperref[sampling:pyshap]{\pyshap}  & \pmark & \pmark & \xmark & \pmark & $\begin{cases}
    T_{L}(\s) & \text{if sampled and $|\s|=|\sbb|$}\\
    T_{L}(\s) + T_{L}(\sbb) & \text{if sampled and $|\s|\neq|\sbb|$}\\
    p_\s              & \text{otherwise}
\end{cases}$
\\
        \hyperref[sampling:pyshaps]{\textcolor{blue}{\pyshaps}} & \pmark & \cmark & \xmark & \pmark & $\begin{cases}
    T_{L}(\s) + T_{L}(\sbb) & \text{if sampled}\\
    p_\s              & \text{otherwise}
\end{cases}$
\\
        \hyperref[sampling:pyshaps_c_kernel]{\textcolor{blue}{\pyshapsckernel}} & \pmark & \cmark & \cmark  & \cmark & $\begin{cases}
    2p_\s \big/ (1-(1-2p_\s)^{L/2}) & \text{if sampled}\\
    p_\s              & \text{otherwise}
\end{cases}$
\\
      \specialrule{.8pt}{0pt}{2pt}
    \end{tabular}
    }
  }
  \vspace{0.5ex}
  \caption{\small{Overview of the sampling strategies along with the weight $w_\s$ they give to a coalition $\s$ in \eqref{eq:ShapleyValuesDefWLSSolution_approx}. 
  Strategies in \textcolor{blue}{blue} font are new methods introduced in the present paper, while the black are existing methods.
  \textbf{Stochastic sampling}: all strategies use stochastic sampling with replacement to form the coalition set $\D$, except the \texttt{PySHAP}-based strategies, which deterministically include coalitions that are expected to be sampled before sampling the rest.
  \textbf{Paired sampling}: whether the strategy samples the paired coalitions $\s$ and $\sbb$ together. 
  \textbf{Equal weights within each coalition size}: whether coalitions of the same size have the same weight.
  \textbf{Weight $\boldsymbol{w}_{\boldsymbol{\mathcal{S}}}$ converges to $\boldsymbol{p}_{\boldsymbol{\mathcal{S}}}$}: whether the weight $w_\s$ converges to $p_\s$ in \eqref{eq:ps_prob} when $N_\text{coal} \rightarrow 2^M$. The non-\texttt{c-kernel} strategies converge in theory, but the convergence is slow in practice due to large weight variability for the sampled coalitions; see \Cref{fig:Sampling}. 
  \textbf{Weight $\boldsymbol{w}_{\boldsymbol{\mathcal{S}}}$ is proportional to}: specifies the proportional weight $w_\s$ given to coalition $\s$ in \eqref{eq:ShapleyValuesDefWLSSolution_approx}. The $T_{L}(\s)$ notation should be read as ``the number of times $\s$ is sampled", while $L$ is the total number of coalitions sampled such that $\D$ contains $N_\text{coal}$ unique coalitions and $p_\s \propto k(M, |\s|)$ as defined in \eqref{eq:ps_prob}.
  }}
  \label{tab:Strategies_summary}
\end{table}

\vspace{-0.5ex}
\subsection{\texttt{Unique}} \vspace{-0.5ex}
\label{sampling:unique}
The \texttt{unique} strategy is the standard established method for obtaining $N_\text{coal}$ unique coalitions, and uses them to estimate Shapley values through \texttt{KernelSHAP}.
It starts by sampling a sequence of $L \geq N_\text{coal}$ coalitions with replacements from the Shapley kernel weight distribution 
\begin{align}
    \label{eq:ps_prob}
    p_{\s} = p(\s) = \frac{k(M, |\s|)}{\sum_{\s \in \pow^*(\M)} k(M, |\s|)} = \frac{k(M, |\s|)}{\sum_{q = 1}^{M-1} k(M, q)\binom{M}{q}},
\end{align}
where $k(M, |\s|)$ is the Shapley kernel weight given in \eqref{eq:ShapleyKernelWeights}. We determine the coalitions using a two-step procedure to avoid listing all $|\pow^*(\M)| = 2^M-2$ coalitions. First, we sample the coalition sizes $|\s| \in \{1, \dots, M - 1\}$ using weighted sampling with replacement, where the weights are $k(M, |\s|)\binom{M}{|\s|}$. Second, we uniformly sample $|\s|$ of the $M$ features without replacement. We repeat this procedure $L$ times until we have $N_\text{coal}$ unique coalitions, which will constitute $\D$. We use the sampling frequencies as the weights $w_\s$ in \eqref{eq:ShapleyValuesDefWLSSolution_approx} since coalitions can been sampled multiple times. That is,
\begin{align}
    w_\s \propto T_{L}(\s),
\end{align}
meaning that $w_\s$ is proportional to the number of times coalition $\s$ is sampled. This strategy was outlined by \cite{aas2019explaining,lundberg2017unified}. 

\subsection{\texttt{Paired}}
\label{sampling:paired}

The \paired\ strategy is a simple, commonly used and established extension of the \unique\ strategy, utilizing the variance reduction technique \textit{paired sampling}\footnote{Also called \textit{antithetic} and \textit{halved} sampling in the literature.} \cite[Ch. 9]{kroese2013handbook} for improved approximation accuracy. \cite{chen2023algorithms,covert2021improving} demonstrates that paired sampling reduces the variance of the Shapley value approximations compared to the \unique\ strategy due to negative correlation between the $v(\s)$ and $v(\sbb)$ values. The \texttt{paired} sampling strategy pairs each sampled coalition $\s$, determined by the \texttt{unique} strategy, with its complement/inverse $\sbb = \M \backslash \s$. Note that inserting $\sbb$ in \eqref{eq:ps_prob} yields $p_\sbb = p_\s$ as $k(M, |\s|) = k(M, M-|\s|)$ in \eqref{eq:ShapleyKernelWeights}.Without loss of generality, let $\s$ be the smaller set of $\s$ and $\sbb$, and $\sbb$ the larger. This will be useful in the remainder of the paper when we talk about coalition size as then $|\s| \leq \floor{M/2}$, where $\floor{\cdot}$ denotes the floor function. Conceptually, we only sample coalitions $\s$ with coalition indices less than or equal to $2^{M-1}$ and then add their paired counterparts $\sbb$ with coalition indices larger than $2^{M-1}$. The probability of sampling $\s$ is then $p_\s + p_\sbb = 2p_\s$, where $p_\s$ is given in \eqref{eq:ps_prob}. However, the doubling is redundant when normalizing the probabilities, as all coalitions have doubled their probability. Note that $N_\text{coal}$ will always be an even number for the \texttt{paired} strategy. The \texttt{paired} strategy also stabilizes the sampling frequencies by ensuring that $\s$ and $\sbb$ always obtain the same weights, as seen in \Cref{fig:Sampling}. The weights are given by
\begin{align}
    w_\s \propto T_{L}(\s) + T_{L}(\sbb),
\end{align}
i.e., proportional to the combined number of times $\s$ and $\sbb$ is sampled.

\subsection{\texttt{Paired C-Kernel}}
\label{sampling:paired_c_kernel}
A byproduct of using the sampling-based procedure for approximating the Shapley values in the \texttt{unique} and \texttt{paired} strategies, is that given the set of uniquely sampled coalitions, the weights $w_\s$ used in $\boldsymbol{W}_\mathcal{D}$ of \eqref{eq:ShapleyValuesDefWLSSolution_approx} are still stochastic. This introduces undesirable variance in the approximation.
Below we derive and present the novel \pairedckernel\ strategy which removes this stochasticity by using a new weighting scheme for the sampled coalitions.

Using the normalized Shapley kernel weights $p_\s$ for approximating Shapley values for all sampled coalitions might appear intuitive, as they correspond to the expected proportions of each coalition, 
and are used in the computations when all coalitions are present. However, this strategy results in biased weights because it fails to consider that we only use the sampled coalitions when approximating the Shapley values (the remaining coalitions technically have weight zero). Hence, we need to condition on $\s$ being among the $N_\text{coal}$ uniquely sampled coalitions. Our \pairedckernel\ strategy addresses this issue, correcting the Shapley kernel weights by conditioning on the coalitions being sampled at least once.

For ease of understanding, we first explain the correction technique for the \texttt{unique} setting before we extend it to the \texttt{paired} setting, which is the one we consider in this article. The \texttt{unique} strategy samples a sequence of $L \geq N_\text{coal}$ coalitions with replacements until $N_\text{coal}$ unique coalitions are obtained. The probability of sampling a coalition $\s$ with size $|\s|$ is $p_\s$ given by \eqref{eq:ps_prob}. 

The stochasticity in the random weights $w_{\s} \propto  T_L(\s)$ will be removed by replacing them with their expected value, conditional on the coalitions being sampled. Since the weights $w_S$ are normalized over the sampled coalitions for numerical stability, it suffices to find $E[T_L(\s)| T_L(\s) \geq 1]$. 
These conditional expectations can then be normalized over the sampled coalitions as before.
By the law of total expectation, we have that
\begin{align*}
E[T_L(\s)] &= E[T_L(\s)| T_L(\s) \geq 1]\operatorname{Pr}(T_L(\s) \geq 1) \\
&\quad + E[T_L(\s)| T_L(\s) = 0]\operatorname{Pr}(T_L(\s) \geq 1) \\
&= E[T_L(\s)| T_L(\s) \geq 1]\operatorname{Pr}(T_L(\s) \geq 1),
\end{align*}
where the latter equality follows since $E[T_L(\s)| T_L(\s) = 0]=0$.
Moreover, since $E[T_L(\s)]=Lp_\s$, and 
${\operatorname{Pr}(T_L(\s) \geq 1)} = 1-{\operatorname{Pr}(T_L(\s) = 0)} = 1 - (1 - p_\s)^L$, we have that
\begin{align}
E[T_L(\s)| T_L(\s) \geq 1] &= \frac{E[T_L(\s)]}{\operatorname{Pr}(T_L(\s) \geq 1)} =\frac{Lp_\s}{1 - (1 - p_\s)^L}.\label{eq:exp-unpaired-c-kernel}
\end{align}
Thus, the corrected, Shapley kernel weights $w_\s$ for the \texttt{unpaired c-kernel} strategy is proportional to the expression in \eqref{eq:exp-unpaired-c-kernel}, i.e.
\begin{align}
w_{\s}  \propto \frac{p_\s}{1 - (1 - p_\s)^L}, \label{eq:ws-unpaired-c-kernel}
\end{align}
which are deterministic given the unique sampled coalitions.

In \Cref{sampling:paired}, we discussed that \texttt{paired} sampling conceptually only samples coalitions $\s$ with coalition indices less than or equal to $2^{M-1}$ and then adds the paired $\sbb$. Thus, the number of sampling steps $L$ is halved, 
$E[T_{L/2}(\s)] = Lp_{\s}/2 + Lp_{\sbb}/2 = Lp_{\sbb} = Lp_{\s}$, i.e.~as before, except that 
${\operatorname{Pr}(T_{L/2}(\s) \geq 1)} = 1-{\operatorname{Pr}(T_{L/2}(\s) = 0)} = 1 - (1 - p_\s - p_{\sbb})^{L/2} = 1 - (1 - 2p_\s)^{L/2}$ takes a slightly different form.
The \pairedckernel\ strategy is therefore the 
\texttt{paired} equivalent of \eqref{eq:ws-unpaired-c-kernel}, which weigh the sampled coalitions by
\begin{align}
    \label{eq:weight_paired_c_kernel}
    w_\s \propto \frac{p_\s}{1 - (1 - 2p_\s)^{L/2}},
\end{align}
where once again, $w_\s$ are normalized over the sampled coalitions.

The \pairedckernel\ strategy assigns equal weight to coalitions within the same coalition size and between paired coalition sizes $|\s|$ and $|\sbb| = M - |\s|$, as illustrated in \Cref{fig:Sampling}. Note that \eqref{eq:weight_paired_c_kernel} is deterministic given the sampled coalitions obtained after $L$ sampling trials. However, if $L$ is not fixed up front, but considered a byproduct of a fixed number of \textit{unique} coalitions $N_\text{coal}$, then there is still some stochasticity from $L$ left in \eqref{eq:weight_paired_c_kernel}. Since sampling of a subset of coalitions is already required when approximating Shapley values, this is mainly of technical character. 
Note however that, when accounting also for the coalition sampling process, \eqref{eq:weight_paired_c_kernel} is clearly stochastic, leading to different Shapley value estimates each time.

\subsection{\pyshap}
\label{sampling:pyshap}
The \texttt{KernelSHAP} method implemented in the popular \href{https://shap.readthedocs.io/en/latest/}{\texttt{SHAP}} Python library \cite{shap_python} samples coalitions using a modification of the \paired\ strategy. It splits the sampling procedure into a deterministic and sampling-based part based on the number of unique coalitions $N_\text{coal}$. Below we introduce this established strategy, which we will denote by \pyshap.

The \pyshap\ strategy deterministically includes all coalitions of size $|\s|$ and $M-|\s|$ if the number of remaining coalitions to sample $N_\text{coal}^*$ exceeds the expected number of coalitions needed to sample all coalitions of these sizes, i.e., $2\binom{M}{|\s|}$. This process bears some resemblance to stratified sampling with strata defined by coalition size. The strategy begins with coalition size one and increments by one until the number of remaining coalitions $N_\text{coal}^*$ becomes insufficient. Denote the coalition size at which the \pyshap\ strategy stops by $P \in \set{1,2,\dots,\ceil{(M+1)/2}}$, where $\ceil{\cdot}$ is the ceiling function. All coalitions of size $|\s| < P$ and $|\s| > M - P$ are then included, and their corresponding weights are assigned according to the normalized Shapley kernel weight in \eqref{eq:ps_prob}. \Cref{fig:Sampling} illustrates the deterministic inclusion of coalitions as a function of $N_\text{coal}$, showing that more coalition sizes are deterministically included as $N_\text{coal}$ increases. 

The remaining $N_\text{coal}^*$ coalitions are sampled with replacements from the non-included coalition sizes, i.e., $|\s| \in \llbracket P, M - P\rrbracket$, following a two-step procedure similar to the one described in \Cref{sampling:unique}. Here, $\llbracket a,b \rrbracket$ denotes the inclusive integer interval between $a$ and $b$. First, a coalition size $|\s| \in \llbracket P, M - P\rrbracket$ is selected with probability proportional to $p_{\s}\binom{M}{|\s|}$, where $p_{\s}$ is the normalized Shapley kernel weight for a coalition $\s$ of size $|\s|$ as given in \eqref{eq:ps_prob}. Second, a coalition $\s$ of size $|\s|$ is sampled among the $\binom{M}{|\s|}$ possible coalitions with uniform probability. The paired coalition $\sbb$ is also included, except when $|\s| = |\sbb|$, a situation that occurs only when $M$ is even. The sampling frequencies are used as weights $w_\s$, but scaled to sum to the remaining normalized Shapley kernel weights, i.e., $1 - 2\sum_{q = 1}^{P-1}p_q\binom{M}{q}$.

For the sake of clarity, consider an example with $M = 10$ and $N_\text{coal} = 100$, as illustrated in \Cref{fig:Sampling}. Then the normalized Shapley kernel weight for a coalition of size one is $p_1 = 0.0196$. This means \pyshap\ includes the $2\binom{10}{1} = 20$ coalitions of size one and nine when $N_\text{coal} \geq \tfrac{1}{0.0196} = 51.02$, which is the case in our example. The remaining number of coalitions is then $N_\text{coal}^* = 80$. The next step is to re-normalize the four remaining Shapley kernel weights, which we denoted by $q_2$, $q_3$, $q_4$, and $q_5$. This yields $q_2 = 0.00404$, $q_3 = 0.00116$, $q_4 = 0.00058$, and $q_5 = 0.00046$, which ensures that $\sum_{l=2}^{\floor{M/2}} q_l\binom{M}{l} = 1$. Thus, \pyshap\ includes all coalitions of size two and eight if $N_\text{coal}^* > \tfrac{1}{0.00404} = 247.5$, which it is not. This implies that $N_\text{coal}$ must be greater than or equal to $20 + 248 = 268$ to include coalition sizes one, two, eight, and nine. Consequently, the remaining $N_\text{coal}^* = 80$ coalitions will be sampled by first selecting the coalition size $|\s|$ with probability $q_\s\binom{M}{|\s|}$, followed by uniformly sampling a coalition $\s$ among the $\binom{M}{|\s|}$ coalitions of size $|\s|$. This procedure is repeated until $N_\text{coal}^*$ unique coalitions are sampled\footnote{Originally, \pyshap\ stops sampling after reaching an upper limit of $4N_\text{coal}^*$ sampled coalitions, including duplicates, which means that the total number of unique coalitions can be less than $N_\text{coal}$. However, we disable this upper limit to ensure a fair comparison with the other strategies, which utilize $N_\text{coal}$ unique coalitions.}, and the corresponding sampling frequencies are used as weights. However, these weights are scaled such that they sum to $1 - 2\binom{M}{1}p_1 = 1- 2\times10\times 0.0196 = 0.608$.

\subsection{\pyshaps}
\label{sampling:pyshaps}
In our examination of the \pyshap\ implementation within version 0.46.0 of \texttt{SHAP} (the latest version as of February 2025), we found that the sampling strategy explicitly does \textit{not} pair coalitions when the coalition size is $|\s| = |\sbb|$, a condition which only applies when $M$ is even. It remains unclear to us whether this is an intentional design choice or an oversight. However, based on our finding, we propose the new, very simple extension \pyshaps\ which always pair coalitions. Notably, \pyshap\ and \pyshaps\ are equivalent when $M$ is odd.

\subsection{\pyshapsckernel}
\label{sampling:pyshaps_c_kernel}
The \pyshapsckernel\ strategy integrates the Shapley kernel weight correction technique described in \Cref{sampling:paired_c_kernel} into \pyshaps's stochastic sampling step in an attempt to achieve better and more stable weights for the sampled coalitions. Specifically, the sampled coalitions receive weights according to \eqref{eq:weight_paired_c_kernel}, but scaled to sum to the remaining normalized Shapley kernel weights, as outlined in \Cref{sampling:pyshap}.

\subsection{Other Sampling Strategies}
\label{sampling:other}
We explored several other sampling strategies, which generally performed worse, more erratically, or were more computationally intensive than the ones discussed above. For completeness, this section provides a brief overview of these strategies.

A strategy to stabilize the weights in the \paired\ strategy is to average the sampling frequencies within each coalition size. Additionally, we also repeated the sampling step $B$ times and used the empirical means of the averaged sampling frequencies to obtain even more stable weights. However, both strategies performed similarly, though slightly and uniformly worse than the \pairedckernel\ strategy. Applying these ideas to the \pyshaps\ strategy yielded improvements but they still fell short compared to the \pyshapsckernel\ strategy.

To remove the stochasticity of $L$ in the \pairedckernel\ and \pyshaps\ \texttt{c-kernel} strategies, we replaced $L$ in \eqref{eq:weight_paired_c_kernel} with $\E[L]$, which is a quantity we can compute using results related to the \textit{coupon collector's problem} \cite{flajolet1992birthday}. Namely, $\E[L] = 2\sum_{q = 0}^{\tilde{N}_\s - 1} (-1)^{\tilde{N}_\s - 1 - q} \binom{2^{M-1} - 1 - q - 1}{2^{M-1} - 1 - \tilde{N}_\s} \sum_{|\mathcal{T}| = q} \frac{1}{1 - P_\mathcal{T}}$,
where $\tilde{N}_\s = N_\text{coal}/2$, $\sum_{|\mathcal{T}| = q}$ represents the sum over all sets of paired coalitions that contain $q$ unique paired coalitions, and $P_\mathcal{T} = \sum_{\s \in \mathcal{T}} 2p_\s$. Computing this quantity is extremely computationally intensive as the number of terms in the inner sum is bounded by $\mathcal{O}(2^{2^M}\!/\sqrt{2^M})$. Additionally, the obtained improvements are negligible and unlikely to affect practical applications. 

Directly using the uncorrected Shapley kernel weights performs poorly, except when $N_\text{coal} \approx 2^M$, where these weights are nearly correct. We also considered a modified version of \pyshaps\ that includes all coalitions of a given size immediately when $N_\text{coal}$ exceeds the number of such coalitions, rather than when they are expected to be sampled. This modification yields good results when $N_\text{coal}$ results in no sampling; however, its erratic behavior for other $N_\text{coal}$ values makes it an unreliable sampling strategy.

\section{Numerical Simulation Studies}
\label{Simulations}
A major problem in evaluating explanation frameworks is that real-world data has no true Shapley value explanations. In this section, we consider two setups where we simulate Gaussian data for which we can compute the true/exact Shapley values $\bphi$ using all $2^M$ coalitions. We then compare how close the approximated Shapley values $\bphi_\D$ are $\bphi$ using the different strategies and based on the coalitions in $\D \subset \pow(\M)$. This paper does not focus on estimating the contribution functions $v(S)$ but on strategies for selecting the coalitions $\s$. Thus, we compute the $v(\s)$ once using the \gaussian\ approach in the \texttt{shapr} \texttt{R}-package \cite{shapr} (or with partly analytical expression for the linear model case in Section \ref{Simulation:lm}), store them, and load the needed $v(\s)$ values for the sampling strategies. 

To elaborate, we generate the training observations and explicands from a multivariate Gaussian distribution $p(\x) = p(\xs, \xsb) = \mathcal{N}_M(\bmu, \bSigma)$, where $\bmu = [\bmu_{\s}, \bmu_{\sbb}]^T$ and $\boldsymbol{\Sigma} = \Big[\begin{smallmatrix} \bSigma_{\s\s} & \bSigma_{\s\sbb} \\ \bSigma_{\sbb\s} & \bSigma_{\sbb\sbb} \end{smallmatrix}\Big]$. The conditional distribution is then $p(\xsb| \xs = \xss) = \mathcal{N}_{|\sbb|} (\bmu_{\sbb|\s}, \bSigma_{\sbb|\s})$, where $\bmu_{\sbb|\s} = \bmu_{\sbb} + \bSigma_{\sbb\s} \bSigma_{\s\s}^{-1}(\xss - \bmu_{\s})$ and $\bSigma_{\sbb|\s} = \bSigma_{\sbb\sbb} - \bSigma_{\sbb\s}\bSigma_{\s\s}^{-1}\bSigma_{\s\sbb}$. 
With an explicit formula for $p(\xsb | \xs = \xss)$, we can estimate \eqref{eq:ContributionFunc} by Monte Carlo integration
\begin{align}
    \label{eq:KerSHAPConditionalFunction}
    v(\mathcal{S})  
    =
    v(\mathcal{S}, \x^*)  
    =
    \E\left[ f(\boldsymbol{x}_{\thickbar{\mathcal{S}}}, \boldsymbol{x}_{\mathcal{S}}) | \boldsymbol{x}_{\mathcal{S}} = \boldsymbol{x}_{\mathcal{S}}^* \right] 
    \approx
    \frac{1}{K} \sum_{k=1}^K f(\boldsymbol{x}_{\thickbar{\mathcal{S}}}^{(k)}, \boldsymbol{x}_{\mathcal{S}}^*) 
    =    
    \hat{v}(\mathcal{S}),
\end{align}
where $\boldsymbol{x}_{\thickbar{\mathcal{S}}}^{(k)} \sim p(\boldsymbol{x}_{\thickbar{\mathcal{S}}} | \boldsymbol{x}_{\mathcal{S}} = \boldsymbol{x}_{\mathcal{S}}^*)$, for $k=1,2,\dots,K$, and $K$ is the number of Monte Carlo samples. The \gaussian\ approach in \texttt{shapr} generate conditional samples $\boldsymbol{x}_{\thickbar{\mathcal{S}}}^{(k)}$ from $p(\boldsymbol{x}_{\thickbar{\mathcal{S}}} | \boldsymbol{x}_{\mathcal{S}} = \boldsymbol{x}_{\mathcal{S}}^*)$, for $k=1,2,\dots,K$ and $\s \in \pow^*(\M)$, and use them in \eqref{eq:KerSHAPConditionalFunction} to accurately estimate $v(\s)$. The parameters $\bmu$ and $\bSigma$ are easily estimated using the sample mean and covariance matrix of the training data, respectively. However, in the present setup, we provide the true parameters to \texttt{shapr} to eliminate the uncertainty in estimating them and to obtain the true Shapley values.

We evaluate the performance of the sampling strategies by computing the averaged mean absolute error ($\operatorname{MAE}$) between the exact ($\bphi$) and approximated ($\bphi_\D$) Shapley values, averaged over $B$ repetitions (with different seeds), $N_\text{explain}$ explicands, and $M$ features. A similar criterion has been used in \cite{aas2019explaining,olsen2023precision,Olsen2022,redelmeier2020}. The $\operatorname{MAE} = \overline{\operatorname{MAE}}_B(\bphi, \bphi_\D)$ is given by
\begin{align}
\label{eq:MAE}
    \operatorname{MAE} 
    = \frac{1}{B} \sum_{b = 1}^{B} \operatorname{MAE}_b = \frac{1}{B} \sum_{b = 1}^{B} \frac{1}{N_\text{explain}} \sum_{i = 1}^{N_\text{explain}} \frac{1}{M} \sum_{j = 1}^{M} |\phi_{i,j} - \phi_{\D_{b}, i,j}|.
\end{align}

\vspace{1ex}
\textbf{Remark:} The sampling step is more expensive for the first three strategies in \Cref{Sampling_strategies} due to the repeated sampling with replacements. More precisely, the sampling procedure is time-consuming when $N_\text{coal} \approx 2^M$ and $M$ is large as the probability of sampling one of the $N_\text{coal} - |\D|$ remaining unique coalitions is minuscule and takes many iterations. For the \paired\ sampling procedure in the $M = 20$-dimensional setting in \Cref{Simulation:lm}, we sample on average $L = 80\,738\,405$ coalitions before we have $2^M = 1\,048\,576$ unique coalitions. In contrast, the \pyshap, \pyshaps, and \pyshapsckernel\ sample only on average $L = 2\,306\,978$ coalitions. The drastic reduction results from deterministically including the most important coalitions, which reduces the pool of coalitions to sample from and aligns their sampling probabilities more closely. However, note that computing the contribution functions is typically the most computationally demanding part of Shapley value computations \cite{olsen2024comparative} and not the sampling of coalitions.

\subsection{XGBoost Model}
\label{Simulation:xgboost}
We let $M=10$ and $N_\text{train} = N_\text{explain} = 1000$. We generate the data according to a multivariate Gaussian distribution $\x \sim \mathcal{N}_M(\boldsymbol{0}, \Sigma)$, where $\Sigma$ is the equi-correlation matrix\footnote{We obtain nearly identical results when $\Sigma_{i,j} = \rho^{\abs{i-j}}$.}. That is, $1$ on the diagonal and $\rho \in \{0, 0.2, 0.5, 0.9\}$ off-diagonal, where one value of $\rho$ is used in each experiment. We generate a response using the formula $y = 2 + 10X_1 + 0.25X_2 - 3X_3 - X_4 + 1.5X_5 -0.5X_6 + 10X_7 + 1.25X_8 + 1.5X_9 - 2X_{10} + X_1X_2 - 2X_3X_5 + 2X_4X_8 -3X_9X_{10} + 3X_1X_3X_7 - X_2X_6X_8 - 2X_3X_8X_{10} + 4X_1X_4X_7X_9$, which contains arbitrary coefficients and significant higher-order interactions effects.

We fit a cross-validated \texttt{xgboost} model \cite{chen2015xgboost} to this regression problem to act as the predictive model $f$. To obtain the true/exact Shapley values $\bphi$, we use Monte Carlo integration with $K = 5000$ Monte Carlo samples for each coalition and explicand. This is done using the \gaussian\ approach in the \texttt{shapr}-package and we repeat the sampling strategies $B = 500$ times as they are stochastic. 

\begin{figure}[!t]
    \centering
    \centerline{\includegraphics[width=1\textwidth]{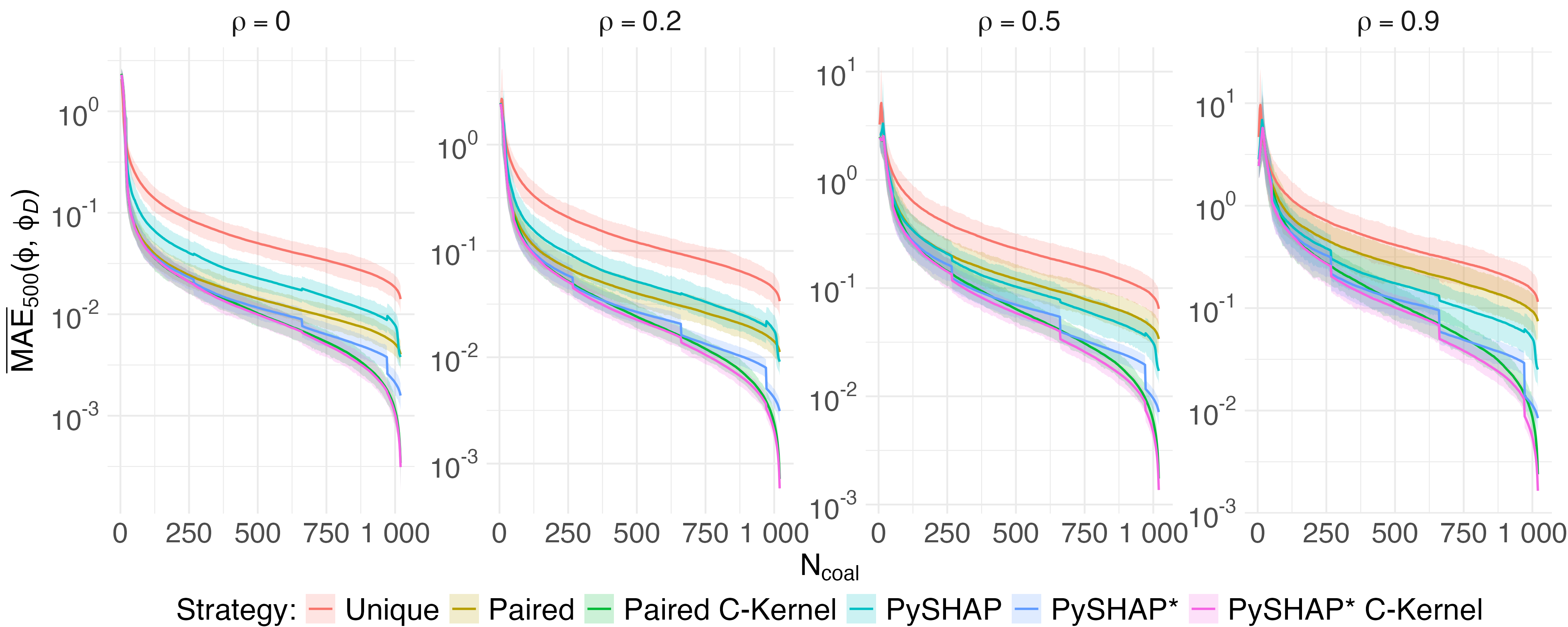}}
    \vspace{-0.75ex}
    \caption{{\small XGBoost experiment: $\operatorname{MAE} = \overline{\operatorname{MAE}}_{500}(\bphi, \bphi_\D)$ for different number of coalitions $N_\text{coal}$ and dependencies levels $\rho$ on log-scale together with $95\%$ confidence bands. 
    }}
    \vspace{-0.5ex}
    \label{fig:XGboost_MAE}
\end{figure}

In \Cref{fig:XGboost_MAE}, we plot the $\operatorname{MAE}$ curves for the different strategies and dependence levels as a function of the number of unique coalitions $N_\text{coal}$. These plots include $95\%$ empirical confidence bands to show the variation in $\operatorname{MAE}_b$ scores across $500$ repeated experiments. The confidence bands become narrower as $N_{\text{coal}}$ increases. While this may not be visually apparent in the figures, where the bands appear uniform, this is just an effect of the logarithmic scale of the y-axis. Note that some strategies do not converge to zero as $N_\text{coal} \rightarrow 2^M$. This is because we plot the $\operatorname{MAE}$ as a function of $N_\text{coal}$ and do not include $N_\text{coal} = 2^M$, where we would naturally use the exact solution \eqref{eq:ShapleyValuesDefWLSSolution} instead of the approximate one \eqref{eq:ShapleyValuesDefWLSSolution_approx}. If we had plotted the $\operatorname{MAE}$ as a function of samples $L$ and continued sampling new coalitions even when $N_\text{coal} = 2^M$, then the $\operatorname{MAE}$ curves for all strategies would eventually converge to zero.

\Cref{fig:XGboost_MAE} reveals five key findings. 
First, a paired sampling procedure in essential as the non-paired \unique\ strategy yields the worst performance. Second, fixing the inconsistent pairing in the \pyshap\ strategy results in distinct improvements with the \pyshaps\ strategy. Third, the $\operatorname{MAE}$ curves for the \pyshap-based strategies exhibit jumps at $N_\text{coal}$ values where additional coalition sizes are deterministically included. Notably, there is a significant improvement at $N_\text{coal} = 268$, reflecting the deterministic inclusion of coalitions of sizes two and eight, as discussed in \Cref{sampling:pyshap}. A possible reason \pyshaps\ sees a larger drop in $\operatorname{MAE}$ than \pyshap\ at these points is that fewer coalitions are in the sampleable set after the jump. As a result, a greater share of the sampleable set is sampled and thereby paired in \pyshaps\, making the pairing more effective. Fourth, the increased performance of \pyshaps\ over \paired\ and \pyshapsckernel\ over \pairedckernel\ suggests that the semi-deterministic sampling strategy outperforms the fully stochastic sampling procedure. Fifth, and most importantly, the \pairedckernel\ and \pyshapsckernel\ strategies, which leverage our Shapley kernel weight correction technique, significantly outperform the other strategies. For low dependence levels $\rho$ and smaller $N_\text{coal}$ values, the \pairedckernel\ and \pyshapsckernel\ strategies perform similarly, with the latter becoming more precise at higher $\rho$ and $N_\text{coal}$ values. However, their confidence bands overlap.

\begin{figure}[!t]
    \centering
    \centerline{\includegraphics[width=1\textwidth]{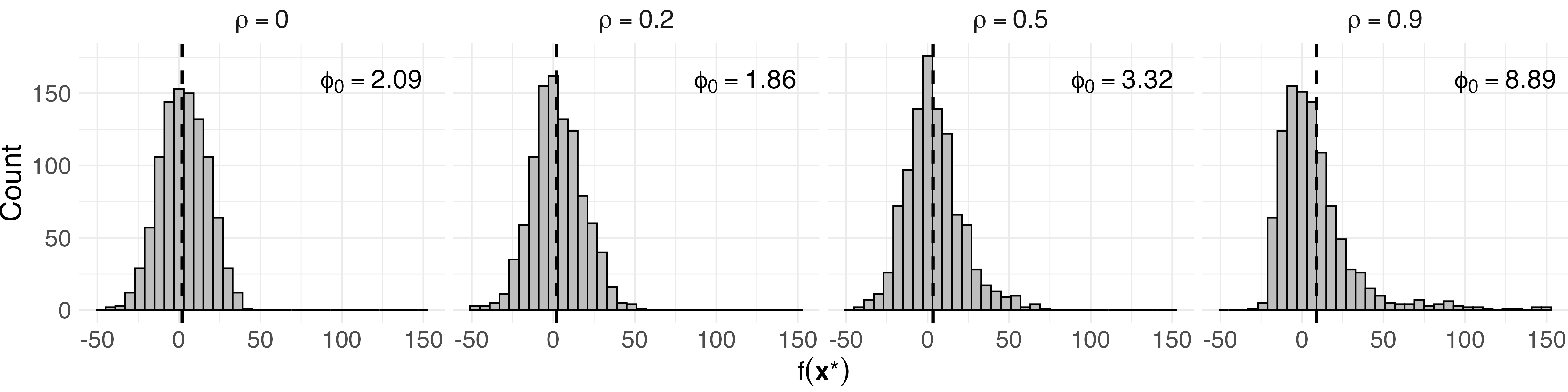}}
    \vspace{-1ex}
    \caption{{\small XGBoost experiment: histograms of the predicted responses $f(\x^*)$ for the $1000$ explicands together with $\phi_0 = \E[f(\x)] = \overline{y}_\text{train}$ for each dependence level.
    }}
    \vspace{-0.5ex}
    \label{fig:XGboost_hist}
\end{figure}

The $\operatorname{MAE}$ values should be interpreted together with $f$ to better understand the scale of the errors. Recall the efficiency axiom which states that $f(\x^*) = \phi_0 + \sum_{j=1}^M \phi_j^*$. That is, the Shapley values explain the difference between $\phi_0$ and the predicted response $f(\x^*)$ for different explicands $\x^*$. In \Cref{fig:XGboost_hist}, we plot histograms of the $f(\x^*)$ values together with the corresponding $\phi_0$ values for the four dependence levels. If the difference between $\phi_0$ and $f(\x^*)$ is small, then the Shapley values are often small too; we illustrate this for the real-world data set in \Cref{sec:real_world_data}. Thus, obtaining a low absolute error is easier for explicands with a predicted response closer to $\phi_0$. 

\begin{figure}[!t]
    \centering
    \centerline{\includegraphics[width=1\textwidth]{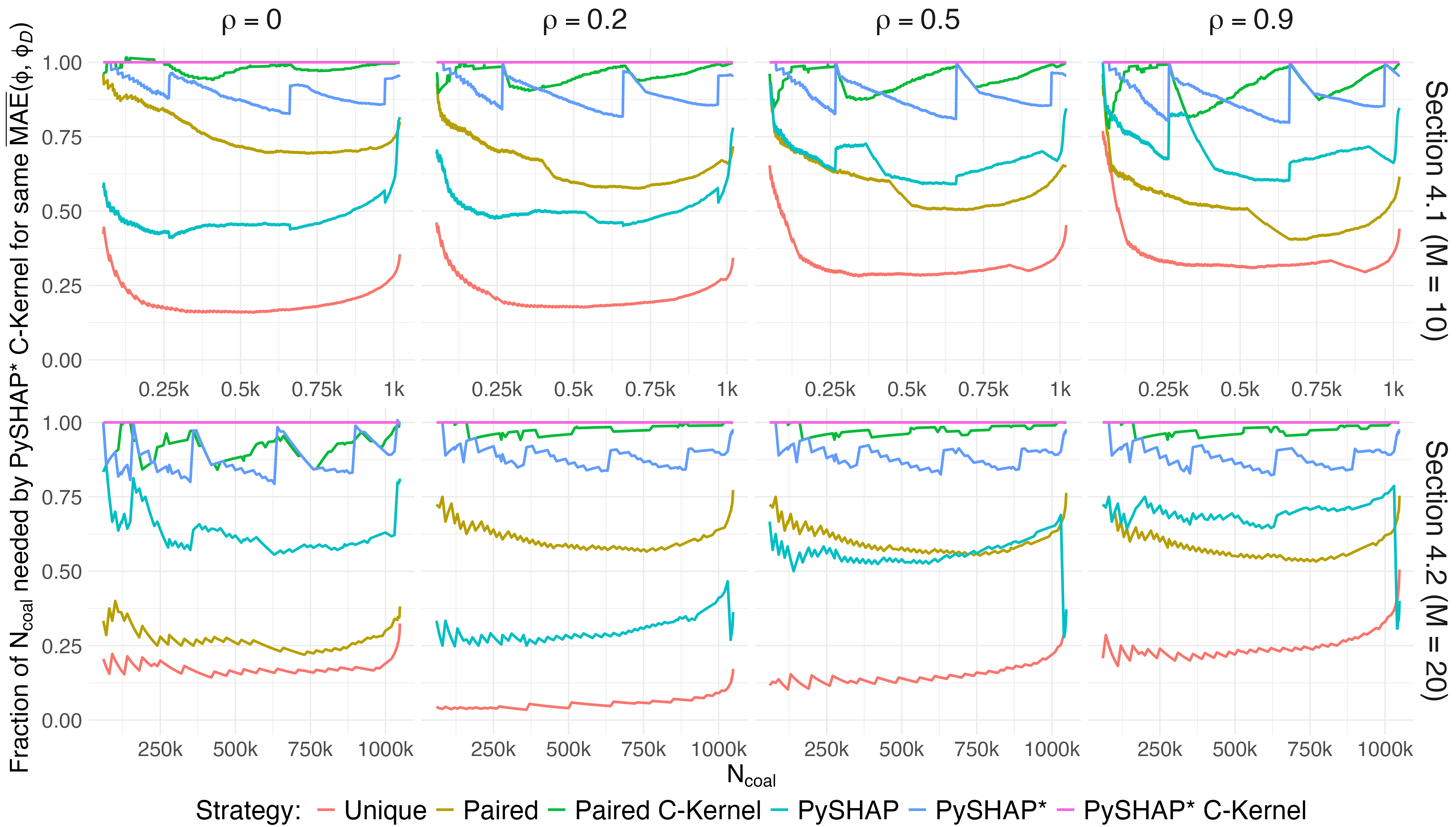}}
    \vspace{-0.75ex}
    \caption{{\small The reduction in $N_\text{coal}$ needed by the \pyshaps\ \texttt{c-kernel} strategy to obtain the same $\operatorname{MAE}$ as the other strategies. E.g., for the experiment in \Cref{Simulation:xgboost} with $\rho = 0.2$ and $N_\text{coal} = 500$, the \pyshapsckernel\ strategy obtains the same $\operatorname{MAE}$ score as the \pyshap\ strategy using only $0.5\times500 = 250$ coalitions, i.e., a $50\%$ reduction. In general, we see similar curves for the different experiments, and all strategies perform significantly worse than the \pyshapsckernel\ strategy. The exception is the \pairedckernel\ strategy in a small region in the top left figure, where it has a fraction above one, indicating that \pyshapsckernel\ requires a larger $N_\text{coal}$ than \pairedckernel.
    }}
    \vspace{-0.5ex}
    \label{fig:fraction_N_S}
\end{figure}

In the simulation studies, we have that $\frac{1}{N_\text{explain}}\sum_{i=1}^{N_\text{explain}}|\phi_0  - f(\x^*)|$ equals $11.54$, $12.27$, $13.02$, $15.98$ for the four dependence levels $\rho \in \{0, 0.2, 0.5, 0.9\}$, respectively. This means that, on average, each of the ten absolute Shapley values are $1.154$, $1.227$, $1.302$, and $1.598$, respectively. Thus, we consider an $\operatorname{MAE}$ between $10^{-1}$ and $10^{-2}$ to yield satisfactory accurate approximated Shapley values. This corresponds to an $N_\text{coal}$ around $100$ to $500$ in the different setups for the \pairedckernel\ and \pyshapsckernel, while the \unique\ and \paired\ strategies need almost all coalitions to obtain the same accuracy for $\rho = 0.9$. In \Cref{fig:fraction_N_S}, we illustrate the reduction in $N_\text{coal}$ when using the \pyshapsckernel\ strategy to achieve the same $\operatorname{MAE}$ scores as the other strategies. In practice, where we do not have access to $\bphi$, it is more applicable to gradually increase $N_\text{coal}$ until $\bphi_\D$ obtains some convergence criterion; see \cite{covert2021improving}.

\subsection{Linear Regression Model}
\label{Simulation:lm}
In the second simulation study, we investigate a higher dimensional setup. However, we are limited by the computational complexity of the true Shapley values growing exponentially with the number of features. Thus, we settle on an $M=20$-dimensional Gaussian data setup with a linear model as the predictive model $f$, as we can then obtain analytical expressions for the contribution function \cite[Appendix B.2]{aas2019explaining}. More precisely, the contribution function in \eqref{eq:ContributionFunc} for linear models with dependent features simplifies to $v(\s) = f(\xsb = \bmu_{\sbb|\s}, \xs = \xss)$, where $\bmu_{\sbb|\s} = \E[\xsb | \xs = \xss] = \bmu_{\sbb} + \bSigma_{\sbb\s} \bSigma_{\s\s}^{-1}(\xss - \bmu_{\s})$ for Gaussian data. This allows us to avoid time-consuming simulations needed in the Monte Carlo integration procedure in \eqref{eq:KerSHAPConditionalFunction} when computing the contribution function values $v(\s)$ for the $2^M = 1\,048\,576$ different coalitions and $N_\text{explain}$ explicands.

We generate $N_\text{train} = 1000$ training observations and $N_\text{explain} = 250$ explicands following a Gaussian distribution $\x \sim \mathcal{N}_M(\boldsymbol{0}, \Sigma_\rho)$, where $\Sigma_\rho = \operatorname{diag}(B_3^\rho, B_4^\rho,$ $B_3^\rho, B_5^\rho, B_2^\rho, B_2^\rho, B_1^\rho)$ is a block diagonal matrix. Here $B_j^\rho$ is an equi-correlation matrix of dimension $j \times j$ and $\rho \in \{0, 0.2, 0.5, 0.9\}$. We generate the responses according to $\y = \boldsymbol{X}\boldsymbol{\beta}$, where arbitrarily $\bbeta = [2, 1, 0.25, -3, -1, 1.5, -0.5, 0.75, 1.25,$ $1.5, -2, 3,-1, -5, 4, -10, 2, 5, -0.5, -1, -2]$ and the first value is the intercept. We fit a linear model to the data and repeat the strategies $B = 150$ times. 

\begin{figure}[!t]
    \centering
    \centerline{\includegraphics[width=1\textwidth]{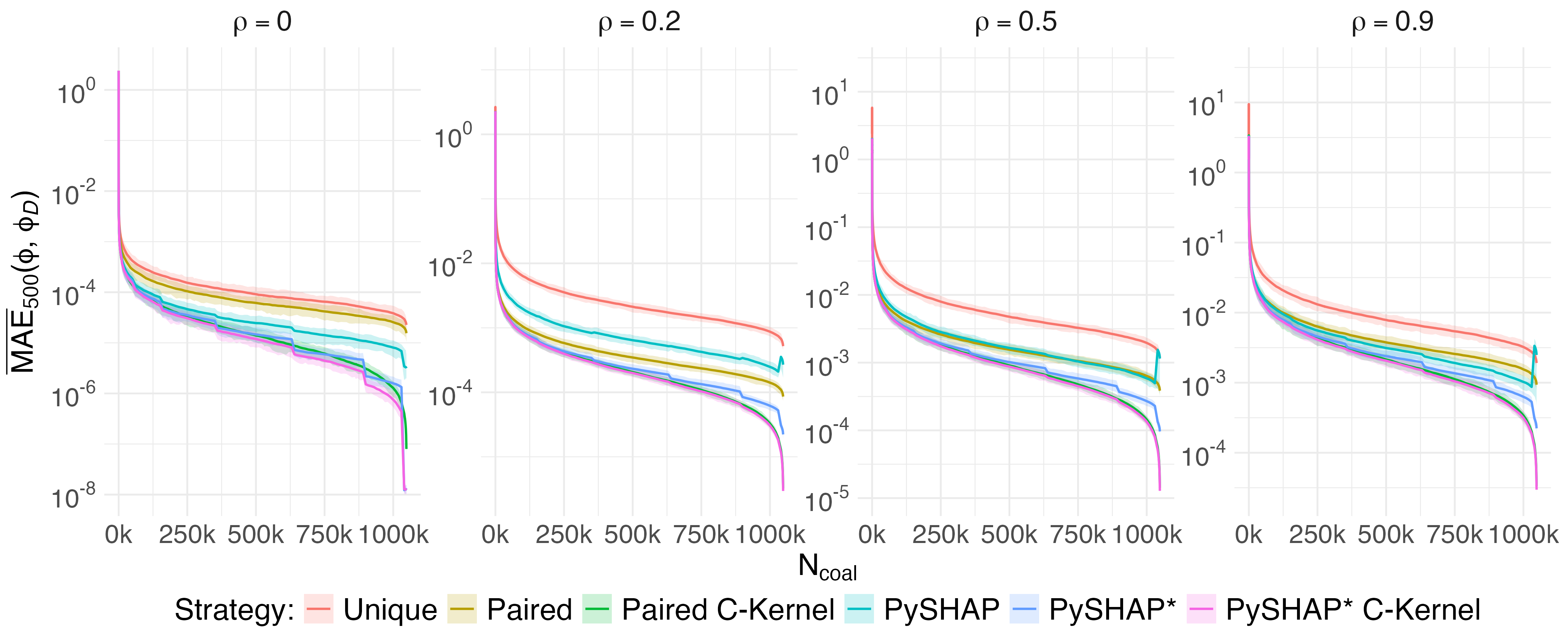}}
        \vspace{-1ex}
    \caption{{\small Linear experiment: $\operatorname{MAE} = \overline{\operatorname{MAE}}_{150}(\bphi, \bphi_\D)$ curves for each strategy and dependence level $\rho$ on log-scale together with $95\%$ confidence bands, which are very narrow.
    }}
        \vspace{-0.5ex}
    \label{fig:Linear_MAE}
\end{figure}

In \Cref{fig:Linear_MAE}, we plot the $\operatorname{MAE}$ curves for each strategy together with $95\%$ empirical confidence bands for the different dependence levels as a function of $N_\text{coal}$. We do not focus on $\rho = 0$ as it is a trivial case where the Shapley values are explicitly given by $\phi_j = \beta_j(x_j^* - \E[x_j])$, for $j = 1,2,\dots, M$ \cite[Appendix B.1]{aas2019explaining}. For all dependence levels, we see that our strategies outperform the established \unique, \paired, and \pyshap\ strategies by a significant margin. For $\rho \geq 0$, the \pyshaps\ strategy performs much better than the three previous strategies, but there is still a notable gap to the \pairedckernel\ and \pyshapsckernel\ strategies. The latter two perform very similarly, with overlapping confidence bands, but the average performance of the \pyshapsckernel\ strategy is slightly better for all values of $N_\text{coal}$. Additionally, the \pyshap\ strategy exhibits erratic behavior when $N_\text{coal} \approx 2^M$, as it samples coalitions only from the innermost coalition size in an unpaired manner. This undesirable behavior is eliminated when we pair the sampled coalitions and use the corrected Shapley kernel weights. We obtain similar results to those in \Cref{fig:Linear_MAE} for other linear simulation experiments not included in this paper with $M \in \{10, 12, 14, 17, 20\}$ and different coefficient values and correlation matrices. 

\begin{figure}[!t]
    \centering
    \centerline{\includegraphics[width=1\textwidth]{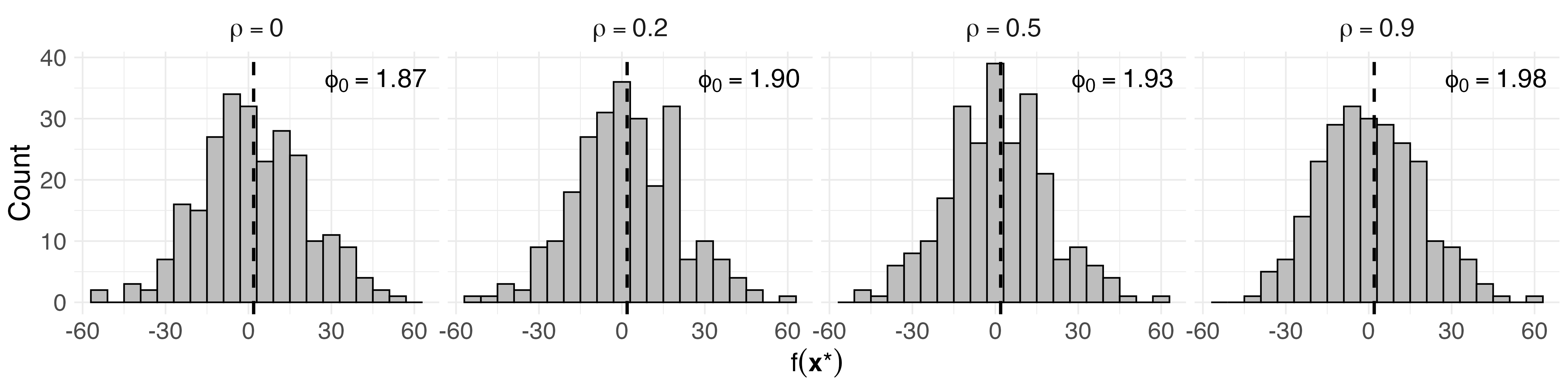}}
    \vspace{-1ex}
    \caption{{\small Linear experiment: histograms of the predicted responses $f(\x^*)$ for the $250$ explicands together with $\phi_0 = \E[f(\x)] = \overline{y}_\text{train}$ for each dependence level.
    }}
    \vspace{-0.5ex}
    \label{fig:Linear_hist}
\end{figure}

In \Cref{fig:Linear_hist}, we plot histograms of the $f(\x^*)$ values together with the corresponding $\phi_0$ values for the four dependence levels. While the corresponding values of $\frac{1}{N_\text{explain}}\sum_{i=1}^{N_\text{explain}}|\phi_0  - f(\x^*)|$ equals $15.23$, $14.89$, $14.56$, and $14.62$, respectively. Which yields an average of $0.76$, $0.74$, $0.73$, and $0.73$ per Shapley value. Thus, we consider an $\operatorname{MAE}$ between $10^{-2}$ and $10^{-2.5}$ to give satisfactory accurate approximated Shapley values. This corresponds to an $N_\text{coal}$ around $2000$ to $300\,000$ in the different setups for the \pairedckernel\ and \pyshapsckernel\ strategies. Note that the number of coalitions $N_\text{coal}$ needed for this precision level increases when the dependence increases.

\section{Real-World Data Experiments}
\label{sec:real_world_data}
\begin{figure}[!t]
    \centering
    \centerline{\includegraphics[width=1\textwidth]{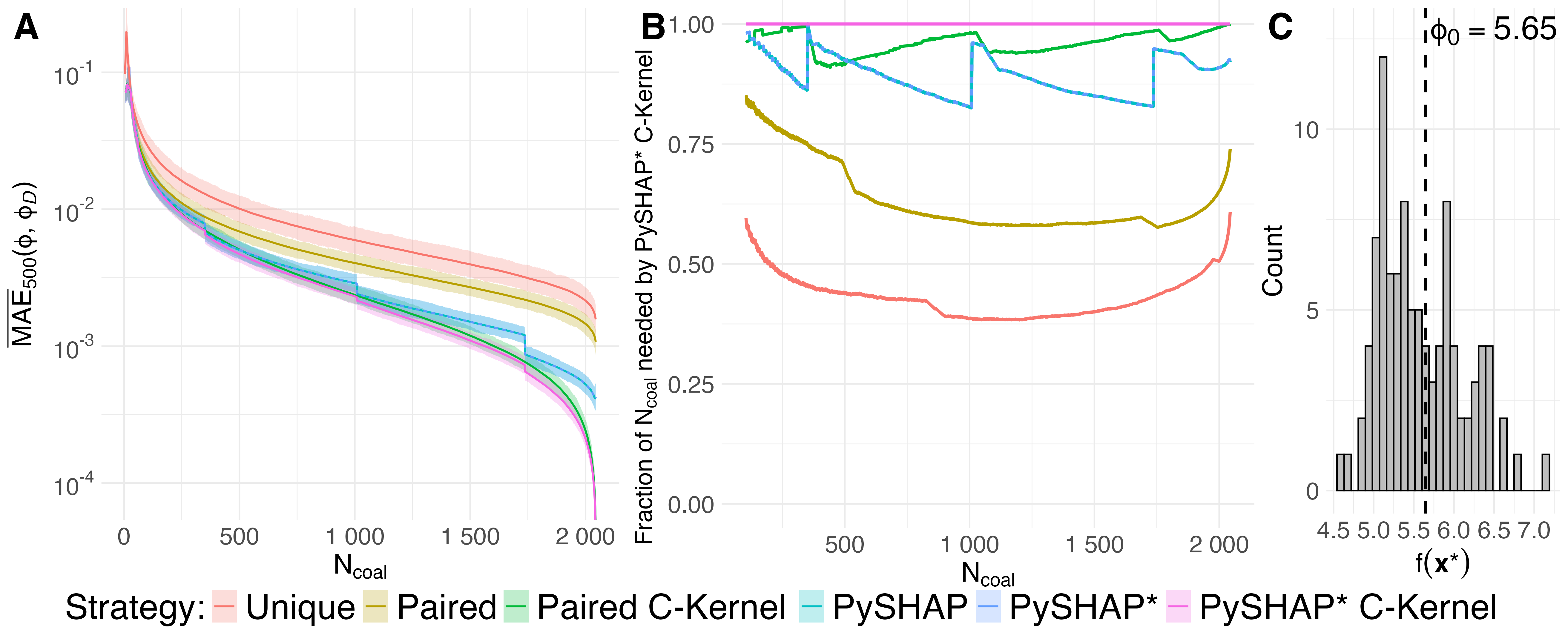}}
    \vspace{-1ex}
    \caption{{\small Red Wine experiment: \textbf{A)} The $\operatorname{MAE} = \overline{\operatorname{MAE}}_{500}(\bphi, \bphi_\D)$ values with $95\%$ confidence bands for different values of $N_\text{coal}$ on log-scale. 
    \textbf{B)} Same type of figure as explained in \Cref{fig:fraction_N_S}.
    \textbf{C)} Histogram of the predicted responses $f(\x^*)$ for the $99$ explicands together with $\phi_0 = \E[f(\x)] = \overline{y}_\text{train}$.}}
    \vspace{-0.5ex}
    \label{fig:RedWineMAE}
\end{figure}

As a real world example, we consider the Red Wine data set from the \href{https://archive.ics.uci.edu}{UCI Machine Learning Repository} to evaluate the sampling strategies on real non-Gaussian data. We fit a cross-validated random forest model\footnote{A \texttt{ranger} model \cite{ranger} with $\texttt{trees} = 200$, $\texttt{mtry} = 4$, and $\texttt{min.node.size} = 3$.} to act as the predictive model. We compute the Shapley values $\bphi$ by estimating the contribution function \eqref{eq:ContributionFunc} values via a random forest regression model approach\footnote{See the  \href{https://norskregnesentral.github.io/shapr/articles/understanding_shapr_regression.html}{regression approach vignette} of the \texttt{shapr} \texttt{R}-package.} with default hyperparameter values: $\texttt{trees} = 500$, $\texttt{mtry} = 3$, and $\texttt{min.node.size} = 5$. \cite{olsen2024comparative} determined this estimation approach to yield the most precise Shapley values for this data set and predictive model. We compare these Shapley values $\bphi$ using all $2^M$ coalitions with the approximated Shapley values $\bphi_\D$ obtained by using only $N_\text{coal} = |\D|$ unique coalitions determined by the different sampling strategies. 

The Red Wine data set contains information about variants of the Portuguese Vinho Verde wine \cite{cortez2009using}. The response is a \tm{quality} value between $0$ and $10$, while the $M = 11$ continuous features are based on physicochemical tests: \tm{fixed acidity}, \tm{volatile acidity}, \tm{citric acid}, \tm{residual sugar}, \tm{chlorides}, \tm{free sulfur dioxide}, total \tm{sulfur dioxide}, \tm{density}, \tm{pH}, \tm{sulphates}, and \tm{alcohol}. For the Red Wine data set, most scatter plots and marginal density functions display structures and marginals far from the Gaussian distribution, as most of the marginals are right-skewed. Many features have no to moderate correlation, with a mean absolute correlation of $0.20$, while the largest absolute correlation is $0.683$ (between \texttt{pH} and \tm{fix\_acid}). The data set contains $1599$ observations, and we split it into a training ($1500$) and a test ($99$) data set. 

In \Cref{fig:RedWineMAE}, we display the $\operatorname{MAE} = \overline{\operatorname{MAE}}_{500}(\bphi, \bphi_\D)$ values with $95\%$ empirical confidence bands. The results are similar to those we obtained for the simulation studies in \Cref{Simulations}. The best strategy across all numbers of coalitions is the \pyshapsckernel\ strategy, with the \pairedckernel\ a close second with partially overlapping confidence bands. The \pyshap\ and \pyshaps\ strategies perform identically as $M$ is odd, but they are significantly outperformed by the two best methods, especially when $N_\text{coal}$ increases. The \texttt{unique} and \texttt{paired} strategies are by far the worst performing methods for all values of $N_\text{coal}$. 

\begin{figure}[!t]
    \centering
    \centerline{\includegraphics[width=1\textwidth]{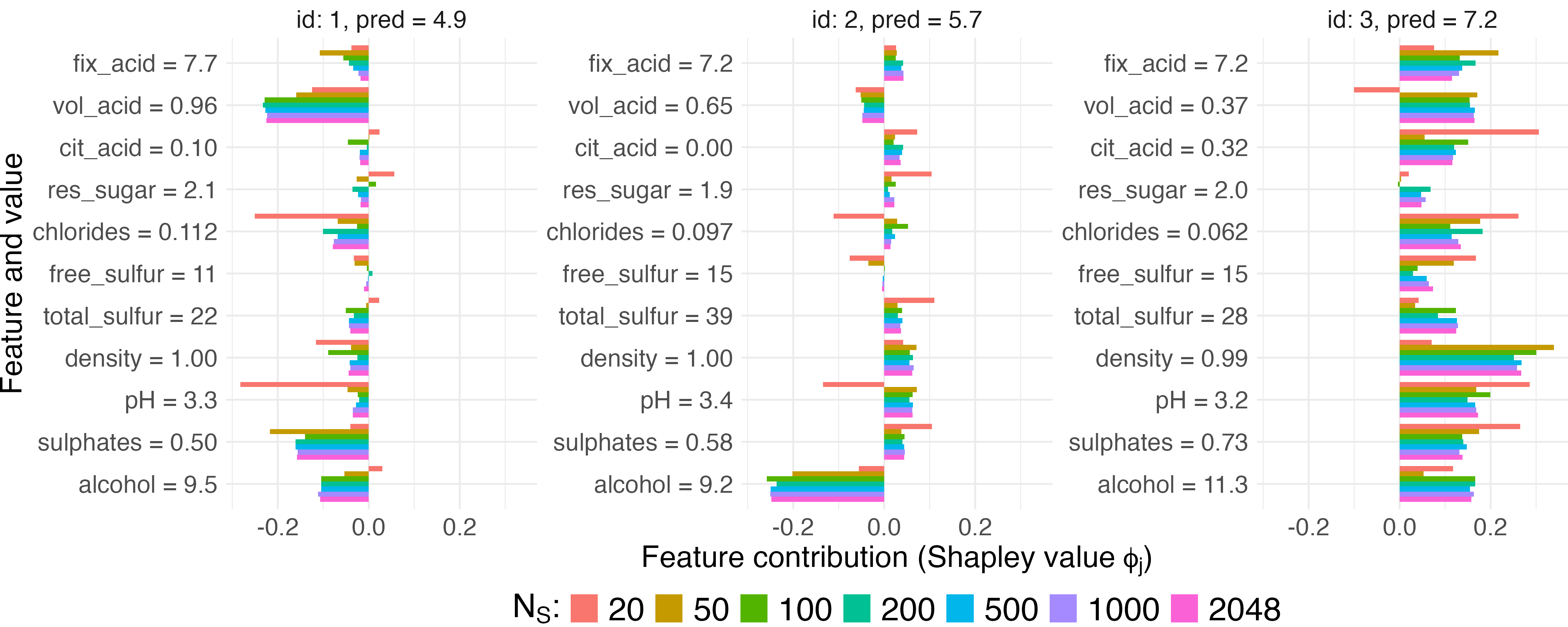}}
    \vspace{-1ex}
    \caption{{\small Red Wine experiment: comparing the Shapley values $\bphi$, obtained using all $2048$ coalitions, with the approximated Shapley values $\bphi_\D$ obtained by the simple \pyshapsckernel\ strategy when increasing the number of unique coalitions $N_\text{coal}$ in the subset $\D$. The approximated Shapley values are close to the exact values when $N_\text{coal}$ larger than $100$ to $200$.}}
        \vspace{-0.5ex}
    \label{fig:RedWineSV}
\end{figure}

As stated in \Cref{Simulations}, the $\operatorname{MAE}$ should be interpreted together with the $f(\x)$ to obtain a proper understanding of the scale of the errors. If the difference between $\phi_0 = 5.65$ and $f(\x^*)$ is small, then the Shapley values are often small too. See explicand two in \Cref{fig:RedWineSV} with $f(\x^*) = 5.7$, where all but one Shapley value is less than $0.05$ in absolute value. Thus, obtaining a low absolute error is easier for explicands with a predicted response closer to $\phi_0$. We have that $\tfrac{1}{N_\text{explain}}\sum_{i=1}^{N_\text{explain}}|\phi_0  - f(\x^*)| = 0.43947$, which means that the $11$ Shapley values move the $\phi_0$ value by $0.44$, or $0.04$ per Shapley value, on average. We consider an $\operatorname{MAE}$ of $10^{-2.5} \approx 0.0032$ and $10^{-3}$ to be low and very low, respectively. For the former, the \pairedckernel\ and \pyshapsckernel\ strategies archive this for $N_\text{coal} \approx 800$, while $N_\text{coal} \approx 1500$ for the very low $\operatorname{MAE}$ value, a precision level not archived by the  \unique\ and \paired\ strategies.

In \Cref{fig:RedWineSV}, we compare the approximated Shapley values $\bphi_\D$ computed by the \pyshapsckernel\ strategy with the Shapley values $\bphi$ using all coalitions for three arbitrarily selected explicands with increasing predicted responses. We see that the magnitude of the Shapley values is larger for predictions further away from $\phi_0$, and the approximated Shapley values are close to the exact values for $N_\text{coal}$ larger than $100$ to $200$. Hence, had this been known in advance, it would have sufficed to use $100$ or $200$ coalitions with the \pyshapsckernel\ strategy, providing large computational savings. Note that, by chance, the $\bphi_\D$ estimates can be close to $\bphi$ for low values of $N_\text{coal}$ for some features and explicands.

\section{Conclusion and Further Work}
\label{Conclusion}
A major drawback of the Shapley value explanation framework is its computational complexity, which grows exponentially with the number of features. The \texttt{KernelSHAP} framework addresses this by approximating Shapley values through a weighted least squares solution, using a sampled subset $\D$ of the $2^M$ coalitions. However, the stochasticity in the weights used in \texttt{KernelSHAP} introduces additional, undesired variance into the approximate Shapley values. We propose the novel \pairedckernel\ strategy for weighing the $N_\text{coal}$ unique elements in $\D$ which is deterministic rather than stochastic, given the unique coalitions. Moreover, our \pyshaps\ strategy makes a correction to the \pyshap\ strategy implemented in the \texttt{SHAP} Python library to ensure consistent pairing of sampled coalitions for all coalition sizes. Finally, the \pyshapsckernel\ strategy combines these two modifications.

Through simulation studies and real-world data examples, we show empirically that our strategies lead to \textit{uniformly} more accurate approximation for conditional Shapley values, while using fewer coalitions than the existing strategies. Specifically, the established \unique, \paired, and \pyshap\ strategies are consistently outperformed by our modified \pyshaps\ strategy, which in turn is outperformed by our innovative \pairedckernel\ and \pyshapsckernel\ strategies. These latter two strategies perform very similarly, with overlapping confidence bands for the $\operatorname{MAE}$ score; however, the \pyshaps\ \texttt{c-kernel} strategy achieves a slightly lower average $\operatorname{MAE}$.

We recommend the \pyshapsckernel\ strategy for its simplicity, consistent high accuracy, and reduced sampling time due to its semi-deterministic inclusion of coalitions. Notably, the \pyshapsckernel\ strategy shows improvement jumps when a new coalition size is deterministically incorporated. The values where these jumps occur can be precomputed, and the results indicate a significant performance boost by choosing a number of coalitions $N_\text{coal}$ slightly higher than these values. This property can be leveraged to develop more effective convergence detection procedures for iterative Shapley value estimation \cite{covert2021improving}.

Additionally, we identify three main areas for future research. First, even though this paper focus on conditional Shapley values, our strategies are also directly applicable for marginal Shapley values, and can be adapted to global Shapley values (e.g., \texttt{SAGE} \cite{covert2020understanding}), Shapley interaction values (e.g., \texttt{KernelSHAP-IQ} \cite{fumagalli2024kernelshap}) and Data Shapley for data valuation \cite{ghorbani2019data}. Similar ideas may also be applied to the alternative approximation approaches of \cite{kolpaczki2024approximating}. Finally, it would be interesting to investigate whether our strategies remain effective if we replace the initial random coalition sampling in \texttt{KernelSHAP} with stratified sampling based on coalition size. Although we have no reason to doubt that our \pairedckernel\ and \pyshapsckernel\ strategies is superior also in these domains, a thorough investigation is needed to confirm this. Shapley value applications where approximate solutions with \texttt{KernelSHAP} is relevant would also be interesting use cases for our \texttt{KernelSHAP} modification. Second, it would be interesting to explore whether our method could be extended and improved by guiding the initial coalition sampling toward more informative subsets for the given model and dataset. This may for instance be achieved by using pilot estimates of $v(\s)$ to identify the most important coalitions and then assigning them greater weight in the sampling process. Taylor approximations as discussed in \cite{goldwasser2024stabilizing} may also serve as a viable alternative in this context. Third, it would be valuable to obtain theoretical insights into the weighting strategies and, if possible, derive theoretical guarantees. For instance, understanding whether the current schemes are optimal, or if better strategies can be found, could be beneficial.

The main computational burden of approximating Shapley values lies in estimating the contribution function values $v(\s)$ rather than sampling and determining the weights of the coalitions $\s$. Thus, the running time is approximately proportional to the number of unique coalitions $N_\text{coal}$ being used. Therefore, our proposed strategies, which reduce $N_\text{coal}$ while maintaining the accuracy of the approximated Shapley values, are very computationally beneficial. Figures \ref{fig:fraction_N_S} and \ref{fig:RedWineMAE} show that, in our experiments, using the \pyshapsckernel\ strategy reduced the number of required coalitions by $50\%$ to $95\%$ compared to the \unique\ strategy, $25\%$ to $50\%$ compared to the \paired\ strategy, and $5\%$ to $50\%$ compared to the \pyshap\ strategy, depending on $N_\text{coal}$ and the dependence level between the features. Importantly, these reductions significantly decrease the computational cost and time for Shapley value explanations, enhancing their feasibility in practical applications.


\begin{credits}
\subsubsection{\ackname} The Norwegian Research Council supported this research through the BigInsight Center for Research-driven Innovation, project number $237718$, Integreat Center of Excellence, project number $332645$, and EU's HORIZON Research and Innovation Programme, project ENFIELD, grant number 101120657.

\appendix
\section*{Appendix}
\label{Appendix}
\section{Implementation Details} \vspace{-1.5ex}
\label{Appendix:Implementation}
The main strategies are implemented in version 1.0.4 of the \texttt{shapr}-package \cite{shapr} while the code for the simulation studies is available at the following GitHub repository: \href{https://github.com/NorskRegnesentral/PaperShapleyValuesImprovingKernelSHAP}{https://github.com/NorskRegnesentral/PaperShapleyValuesImprovingKernelSHAP}.
\newline The implementation of the \pyshap-based strategies are based on the \href{https://github.com/shap/shap/blob/master/shap/explainers/_kernel.py}{\texttt{\textunderscore kernel.py}} file in version 0.46 of the \href{https://shap.readthedocs.io/en/latest/}{\texttt{SHAP}} Python library \cite{shap_python}. To generate the true contribution functions and Shapley values in \Cref{Simulation:xgboost}, we have used the \texttt{gaussian} approach in \shapr\ with $K = 5000$ Monte Carlo samples from the true Gaussian distribution. While we used explicit formulas for the contribution function in \Cref{Simulation:lm}. In \Cref{sec:real_world_data}, we used the \texttt{random forest separate regression} approach introduced in \cite{olsen2024comparative} to compute $v(\mathcal{S})$. 

\end{credits}
\footnotesize
\bibliographystyle{splncs04}
\bibliography{mybib.bib}
\end{document}